\documentclass{article}

\usepackage[preprint]{neurips_2026}

\usepackage[utf8]{inputenc} 
\usepackage[T1]{fontenc}    
\usepackage{hyperref}       
\usepackage{url}            
\usepackage{booktabs}       
\usepackage{multicol}
\usepackage{float}
\usepackage{multirow}
\usepackage{threeparttable}
\usepackage{amsfonts, amsmath, amssymb}       
\usepackage{nicefrac}    
\usepackage{wrapfig}
\usepackage[ruled,linesnumbered]{algorithm2e}
\usepackage{graphicx}

\usepackage{microtype}      
\usepackage{xcolor}         
\usepackage{wrapfig}
\usepackage{amsmath}
\usepackage{amsthm}
\usepackage{amssymb}

\usepackage{caption}
\title{Robust Graph Clustering Network for Multiple Missing Data}

\author{%
  Keyuan Qiu \\
  Northeastern University
  \And
  Renda Han \\
  Northeastern University
  \And
  Zhen Tang \\
  Northeastern University
  \AND
  Qiang He\thanks{indicates the corresponding author.} \\
  Northeastern University
  \And
  Xingwei Wang \\
  Northeastern University
  \And
  Wenxin Zhang \\
  University of Chinese \\
  Academy of Sciences
  \AND
  Guangzhen Yao \\
  Northeast Normal \\
  University
  \And
  Junxin Chen \\
  Dalian University \\
  of Technology
  \And
  Qingjian Ni \\
  Southeast University
}

\begin{document}

\maketitle

\begin{abstract}
Clustering on graphs where both node attributes and structural links are partially missing remains a challenging task.
Existing methods typically rely on imputation-then-clustering on single-view missingness incomplete graphs, which are vulnerable to cross-view error propagation and cluster-boundary blurring under simultaneous attribute and structure missingness. To address these limitations, we propose a \textbf{R}obust \textbf{G}raph \textbf{C}lustering \textbf{N}etwork for Multiple Missing Data (RGCN), which is designed to handle simultaneous node attribute and graph structure incompleteness. RGCN introduces three key innovations: First, we design a view-decoupled dual-branch imputation to mitigate interference and enable mutual enhancement in recovering missing data. Second, we employ a multi-hyperspherical mixture prior to enhance intra-cluster compactness and inter-cluster separability on a directional latent manifold. Third, a boundary-aware contrastive enhancement objective mitigates the blurring of clusters caused by imputation bias. Extensive experiments on real-world datasets demonstrate that RGCN consistently outperforms state-of-the-art baselines under various missing patterns.

\end{abstract}

\section{Introduction}
Graph clustering \cite{Bo2020structuraldeep, Tu2021deepfusion, liang2024asurvey, liu2025adaptivefeature, liu2025multiviewtemporal, liu2026causallyawareattribute, devvrit2022s3gc} aims to partition nodes in an attributed graph \cite{liang2025fromconcrete, liu2025multiviewtemporal} into semantically coherent groups, serving as a fundamental task in graph machine learning \cite{wang2025elevatingknowledgeenhanced, liu2024adaptivemultichannel,liang2024hawkesenhancedspatialtemporal}. 
It has widespread applications in social network analysis \cite{hamilton2017inductive}, recommendation systems \cite{ying2018graph}, and bioinformatics \cite{hu2025pagm, hu2025single}. Recent advances in Graph Neural Networks (GNNs) and self-contrastive learning have significantly improved the performance of clustering algorithms on complete-graph data. 

However, real-world graph data is often far from complete. Due to privacy constraints, sensor failures, or the continuous arrival of new entities, graph data frequently suffers from simultaneous missingness in both node attributes and structural links \cite{liu2026causallyawareattribute}. To address this issue, some existing work has focused on attribute-missing graph clustering \cite{tu2024attributemissinggraph, tu2022initializingthen, guan2025prototype, hu2025scalable, hu2024reliable}, where node features are partially available while the topology is assumed to be complete. 
These methods typically leverage graph smoothness or prototype-guided propagation to infer missing attributes from observed neighbors for representation learning, followed by clustering in the completed feature space. In contrast, existing studies on structure-missing learning primarily focus on link prediction \cite{zhang2018link} or graph completion \cite{kim2025accurate}, typically under the assumption that node attributes are fully observed. These approaches generally rely on label supervision and complete node attributes. More recently, a method \cite{chen2026towards} has been proposed to overcome the above issues. However, their objective is to learn a universal and detection-friendly graph representation, making it unsuitable for unsupervised graph clustering.
Moreover, SMGCN \cite{hu2026structure} is among the first clustering methods for missing structures, but it still has two limitations: (i) it operates at the graph-level rather than explicitly capturing node-level clustering signals, and (ii) it still assumes complete node attributes and thus cannot handle multiple data missingness simultaneously.

Based on the above analysis, an intuitive idea is to design a unified framework that simultaneously imputes missing attributes and edges. However, this paradigm faces two key challenges: (1) attribute or structure imputation is performed under incomplete information from the other view, which severely undermines reliability; (2) during cross-view interaction, errors can propagate between views, leading to accumulated degradation. Inspired by \cite{wang2026federated}, to address this issue, we adopt a view-decoupled reconstruction design, where node attribute imputation and structure recovery are parameterized by separate modules and optimized with branch-specific objectives.
To tackle the second issue, we further impose a multi-hyperspherical prior during representation learning to maintain clear inter-cluster margins, which is integrated into the alternating updates, constraining error propagation across views. 

Therefore, we propose a novel graph clustering method to address the multiple data missingness. First, we employ a dual-pathway decoupled imputation method to reconstruct node attributes and graph structure in a diffusion manner, optimizing both alternately. Subsequently, the imputed node attributes are projected onto a unit multiple hyperspherical latent space to obtain learnable cluster centers. Cluster assignment is driven by maximizing the cosine similarity between node embeddings and their respective cluster centers, promoting intra-cluster compactness and inter-cluster separability. Concurrently, a boundary-aware contrastive learning between clusters to further sharpen cluster demarcation is jointly optimized, enabling end-to-end cluster inference. 
Even under severe dual-view incompleteness, the network effectively reconstructs the underlying data manifold while maintaining well-separated cluster structures. Our contribution can be summarized as follows:

\begin{itemize}
    \item \textbf{New Perspective} To the best of our knowledge, this is the first attempt to address simultaneous missing node attributes and graph structure in graph clustering, which poses a more challenging setting than single-type missing scenarios.
    \item \textbf{New Method} We propose a novel graph clustering framework, termed RGCN, which handles simultaneous attribute and structure missingness. RGCN mitigates cross-view interference through a view-decoupled dual-branch reconstruction. Moreover, RGCN enhances latent separability via a multi-hyperspherical mixture prior and contrastive boundary enhancement.
    \item \textbf{Excellent Performance} Extensive experiments on six real-world datasets demonstrate that RGCN consistently outperforms SOTA baselines under various complex missing patterns, exhibiting notable reliability.
\end{itemize}

\section{Related Work}
\label{app:sec:relatedWork}
\subsection{Graph Learning with Missing Attributes and Structures}

Learning on graphs with incomplete node attributes or missing structural connections remains a fundamental challenge \cite{fatemi2021slaps}, with most existing studies confined to a single-view setting.
For missing node attributes, methods such as ITR \cite{tu2022initializingthen} adopt a two-step paradigm that first performs imputation and then conducts the downstream task, typically leveraging graph smoothness priors to recover missing features.
However, such methods overlook the interplay between attribute imputation and downstream tasks. AMGC \cite{tu2024attributemissinggraph} addresses this limitation by jointly optimizing imputation and clustering in a unified framework. Subsequently, PMAGC~\cite{guan2025prototype} further utilizes multi-view alignment and the prototypes to drive improving the clustering performance. Similarly, AMMGC~\cite{zhao2025attribute} introduces a dual structure consistency across views to refine missing node attributes. More recently, CMVND \cite{hu2025scalable} utilizes neighborhood differentiation to further optimize the imputation backbone and generate reliable guidance for node attribute recovery. For the missing graph structure, most studies \cite{kim2025accurate} mainly address link prediction or graph completion under the assumption of fully observed node attributes, and are not directly tailored for clustering tasks. Consequently, these methods often fail to explicitly capture node-level clustering signals during the topology recovery process. 

In real-world scenarios, attribute and structural missingness often co-occur. Existing work such as M$^2$V-UGAD~\cite{chen2026towards} has begun to explore this setting, but it is designed for anomaly detection and is not aligned with graph clustering objectives. Directly applying separate imputation methods in an “impute-then-cluster” manner can lead to coupled errors between attributes and structure, degrading representation quality.

\begin{figure}[H]
    \centering
\includegraphics[width=1\linewidth]{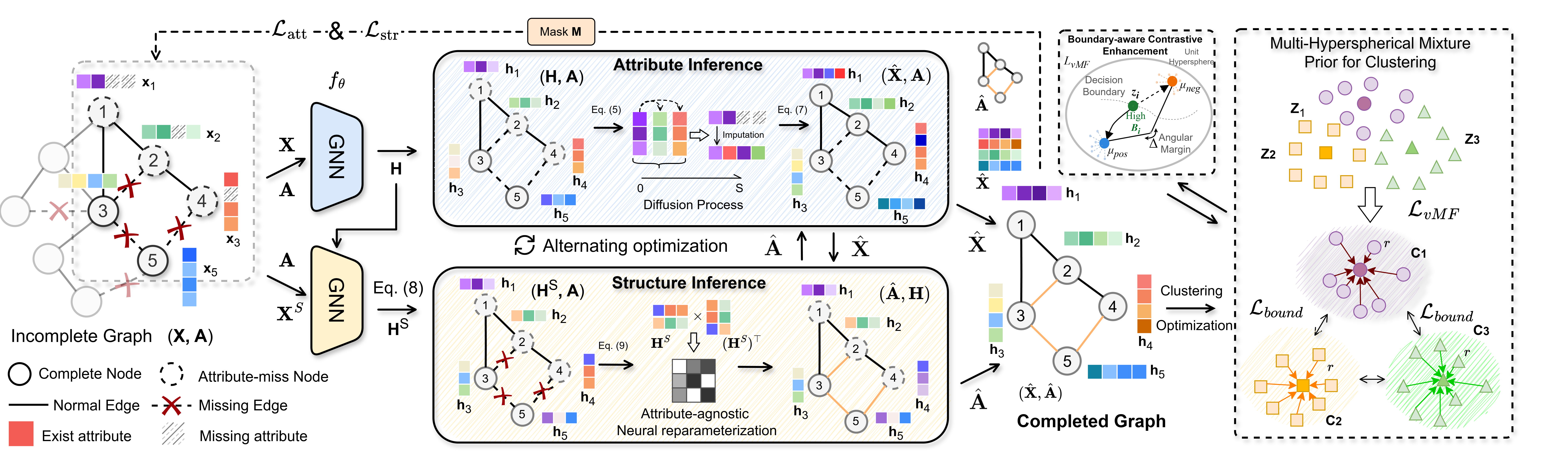}
    \caption{RGCN first extracts node attributes through a dual-branch network. Subsequently, it alternates between diffusion-based attribute imputation and structure inference to progressively recover missing values. Finally, clustering optimization is conducted under a multi-hyperspherical mixture prior, yielding the final clustering results.}
    \label{fig:framework}
\end{figure}

\subsection{Deep Graph Clustering}

DGC has emerged as a potent paradigm, coupling GNN representation with diverse clustering objectives.
While earlier methods like SDCN \cite{Bo2020structuraldeep} and DFCN \cite{Tu2021deepfusion} employ auto-encoder reconstructions for joint optimization, recent models such as SCGC \cite{liu2023simple} and DCLN \cite{peng2023dualcontrastive} prioritize maximizing mutual information via contrastive learning. However, these basic DGC frameworks overly rely on complete and clean graph inputs, which makes them very fragile when dealing with missing data in reality, vulnerable to structural noise interference and causing errors to continuously spread. Specifically, existing incomplete graph clustering (IGC) efforts, including ITR \cite{tu2022initializingthen} and AMGC \cite{tu2024attributemissinggraph}, primarily focus on attribute-only missingness by leveraging graph smoothness priors to propagate observed features. While effective for single-type missingness, these models fail to address simultaneous attribute and structural gaps, where cross-view error propagation severely undermines imputation reliability and degrades representation quality.

To tackle such simultaneous missingness, an intuitive solution is to recover the incomplete data manifold before applying standard DGC paradigms. However, unsupervised GNNs (e.g., GraphMAE \cite{hou2022graphmae}) are vulnerable to imputation bias and structural noise, triggering 'cluster-boundary blurring' via global-mean drift. This situation is often exacerbated by the excessive smoothing of deep architectures, which severely compresses the latent space and erases the key structural features required to effectively distinguish different clusters \cite{liu2020towards, zhao2020pairnorm, feng2020graph}.

\section{Methodology}

In this section, we introduce the \textbf{R}obust \textbf{G}raph \textbf{C}lustering \textbf{N}etwork for Multiple Missing Data (RGCN) in detail. The specific structure of the framework is shown in the Fig. \ref{fig:framework}. To provide a clear architectural roadmap before delving into the mathematical specifics, the overall training pipeline is summarized in Algorithm \ref{alg:trans_concise}. The framework structurally decouples the imputation and clustering processes into three progressive phases: dual-branch value imputation, multi-hyperspherical prior extraction, and contrastive boundary-enhancement.

\subsection{Notations}
Let $\mathcal{G}_{obs}=(\mathcal{V},\mathbf{X}_{obs},\mathbf{A}_{obs})$ denote the observed incomplete attributed graph, where $\mathbf{X}_{obs}$ and $\mathbf{A}_{obs}$ are obtained from the complete graph $\mathcal{G}=(\mathcal{V},\mathbf{X},\mathbf{A})$ under attribute and structure masks, respectively. The reconstructed attributes and structure are denoted by $\hat{\mathbf{X}}$ and $\hat{\mathbf{A}}$. A summary of key notations is provided in Appendix~\ref{app:Notations}.

\subsection{Dual Branch Value Imputation}

To reduce cross-view error propagation under simultaneous attribute and structure missingness, we propose a view-decoupled dual-branch imputation strategy, where attribute completion and structure recovery are separately parameterized and optimized with branch-specific objectives. Specifically, the attribute branch reconstructs missing node attributes through diffusion-based inference, while the structure branch recovers missing links through attribute-agnostic neural reparameterization. The two branches are jointly refined through alternating optimization.

\begin{wrapfigure}[26]{r}{0.6\textwidth}
\vspace{-10pt}
\centering
\begin{minipage}{0.95\linewidth}
\begin{algorithm}[H]
\caption{Training Procedure of RGCN}
\label{alg:trans_concise}
\LinesNumbered
\KwIn{Incomplete graph $\mathcal{G}_{obs}=(\mathcal{V},\mathbf{X}_{obs},\mathbf{A}_{obs})$; masks $\mathbf{M}_X,\mathbf{M}_A$; cluster number $K$; max epochs $E$.}
\KwOut{Clustering results $\mathcal{R}$.}

Initialize parameters $\Theta$, base concentration $\kappa_0$, and prototypes $\{\boldsymbol{\mu}_k\}_{k=1}^{K}$\;

\For{$epoch=1$ \KwTo $E$}{
    Impute attributes via denoising topology-decoupled diffusion and obtain $\hat{\mathbf{X}}$\;
    
    Recover links via attribute-agnostic low-rank structural reparameterization and obtain $\hat{\mathbf{A}}$\;
    
    Extract embeddings $\mathbf{Z}=f_{\mathcal{G}}(\hat{\mathbf{X}},\hat{\mathbf{A}})$ and project them onto $\mathbb{S}^{d-1}$\;
    
    Update $\Theta$ by minimizing $\mathcal{L}_{total}=\mathcal{L}_{att}+\mathcal{L}_{str}+\lambda_1\mathcal{L}_{vMF}+\lambda_2\mathcal{L}_{bound}$\;
    
    Compute boundary sensitivity $B_i$ and concentration $\kappa_i$ for each node\;
    
    Update assignments $q_{ik}$ and prototypes $\boldsymbol{\mu}_k$ by E/M-style clustering refinement\;
}

\Return $\mathcal{R}=\{\arg\max_k q_{ik}\}_{i=1}^{n}$\;
\end{algorithm}
\end{minipage}
\end{wrapfigure}

\paragraph{Node Attribute Inference}
To recover missing attributes under incomplete graph structures, we introduce a topology-decoupled feature diffusion module. 
Instead of directly propagating attributes along the unreliable observed adjacency matrix $\mathbf{A}_{obs}$, we perform diffusion in a dynamically constructed feature-affinity space.

To obtain self-supervised signals for attribute imputation, we adopt a denoising masking strategy. 
A denoising mask $\mathbf{M}_D\in\{0,1\}^{n\times d}$ is sampled from the observed entries, satisfying $\mathbf{M}_D\leq \mathbf{M}_X$, and the corrupted input is defined as
\begin{equation}
\tilde{\mathbf{X}}_{obs}=(\mathbf{M}_X-\mathbf{M}_D)\odot\mathbf{X}_{obs}.
\end{equation}
The initial node representation is obtained by
\begin{equation}
\mathbf{H}^{(0)}=f_{\theta}(\tilde{\mathbf{X}}_{obs},\mathbf{A}_{obs}).
\end{equation}

Given $\mathbf{H}^{(0)}$, we construct a feature-affinity kernel $\mathbf{K}^{(s)}$ from latent node representations and define the corresponding normalized feature-space Laplacian as
\begin{equation}
\boldsymbol{\Delta}_{\mathcal{X}}(s)=\mathbf{I}-\mathbf{D}^{-\frac{1}{2}}\mathbf{K}^{(s)}\mathbf{D}^{-\frac{1}{2}}.
\end{equation}
The topology-decoupled diffusion process is then formulated as
\begin{equation}
\frac{\partial \mathbf{H}(s)}{\partial s}=-\boldsymbol{\Delta}_{\mathcal{X}}(s)\mathbf{H}(s)\mathbf{W}_v,
\end{equation}
where $\mathbf{W}_v$ is a learnable propagation matrix. 
Since $\boldsymbol{\Delta}_{\mathcal{X}}(s)$ is defined by feature affinity rather than directly by $\mathbf{A}_{obs}$, the diffusion avoids explicitly propagating attributes through missing or unreliable graph links. Integrating over a diffusion depth $S$ yields the diffusion-enhanced latent representation:
\begin{equation}
\mathbf{H}^{(S)}
=
\mathbf{H}^{(0)}
+
\int_0^S
\big(
-
\boldsymbol{\Delta}_{\mathcal{X}}(s)
\mathbf{H}(s)
\mathbf{W}_v
\big)ds .
\end{equation}

We further use an attribute decoder $g_{\phi}(\cdot)$ to map the diffusion-enhanced representations back to the original feature space
\begin{math}
\mathbf{X}_{pred}
=
g_{\phi}(\mathbf{H}^{(S)}),
\end{math}
where $g_{\phi}:\mathbb{R}^{d_h}\rightarrow\mathbb{R}^{d_x}$ denotes a learnable
feature decoder, $d_h$ is the latent representation dimension, and $d_x$ is the original
attribute dimension.
The attribute reconstruction loss is computed only on the deliberately masked observed
entries:
\begin{equation}
\mathcal{L}_{att}
=
\big\|
\mathbf{M}_D
\odot
\big(
\mathbf{X}_{pred}
-
\mathbf{X}_{obs}
\big)
\big\|_F^2 .
\end{equation}
After training, the final imputed attribute matrix is obtained by preserving the originally
observed attributes and filling only the truly missing entries with the predicted values:

\begin{equation}
\hat{\mathbf{X}} = \mathbf{M}_X \odot \mathbf{X}_{obs} + (1-\mathbf{M}_X)\odot \mathbf{X}_{pred}.
\end{equation}

This denoising objective explicitly supervises attribute diffusion and decoding by reconstructing observed entries that are masked during training. The final imputation rule preserves all observed attributes and completes only the truly missing ones using the learned diffusion-based predictor.

\paragraph{Graph Link Inference}
To recover missing graph links without propagating biased attribute imputation into structure recovery, we introduce an attribute-agnostic structural reparameterization module. 
Specifically, we first construct a random-walk-based structural encoding $\mathbf{X}^{S}$ from the observed incomplete adjacency matrix $\mathbf{A}_{obs}$, and obtain the structure-aware representation by
\begin{equation}
\mathbf{H}^{S}=GNN_{\mathrm{struct}}
\big(\mathbf{A}_{obs}, \mathbf{X}^{S}\big).
\end{equation}
Then, a learnable reparameterization function maps $\mathbf{H}^{S}$ into a low-rank structural embedding:
\begin{math}
\mathbf{E}_s=f_{\mathrm{rep}}\big(\mathbf{H}^{S}\big).
\end{math}
Since $\mathbf{E}_s$ is inferred from structural information rather than reconstructed attributes, the structure branch remains separately parameterized from the attribute branch. 
The recovered adjacency matrix is generated by a low-rank inner-product decoder:
\begin{equation}
\hat{\mathbf{A}}=\operatorname{Sigmoid}
\big(\mathbf{E}_s\mathbf{E}_s^{\top}\big).
\end{equation}
The structure reconstruction loss is defined as
\begin{equation}
\mathcal{L}_{str}
=
\frac{1}{2}
\left\|
\mathbf{M}_A \odot
\left(
\hat{\mathbf{A}}-\mathbf{A}_{obs}
\right)
\right\|_F^2
+
\frac{\eta}{2}
\mathrm{Tr}
\left(
\hat{\mathbf{A}}^{\top}
\mathbf{L}_{obs}
\hat{\mathbf{A}}
\right),
\end{equation}

where $\mathbf{L}_{obs}$ is the graph Laplacian computed from the observed adjacency matrix $\mathbf{A}_{obs}$, and $\eta$ controls the strength of topology smoothness regularization. The first term preserves the observed structural entries, while the second term encourages the recovered adjacency to be smooth with respect to the reliable observed topology.

\subsection{Hyperspherical vMF Clustering with Boundary Enhancement}
In this section, our objective is to leverage prior knowledge to enable high-quality clustering. However, conventional impute-then-clustering paradigms are prone to imputation bias, which causes representations to collapse toward the Euclidean mean and results in blurred cluster boundaries. Here, we project representations onto a hyperspherical manifold and introduce a Multi-Hyperspherical Mixture Prior, allowing clustering to be guided by directional consistency, enhancing cluster separability.

Given the completed graph $(\hat{\mathbf{X}},\hat{\mathbf{A}})$, we extract node representations 
$\mathbf{Z}=f_{\mathcal{G}}(\hat{\mathbf{X}},\hat{\mathbf{A}})$ and project them onto the unit hypersphere by $\ell_2$ normalization. 
This removes magnitude-induced bias and allows clustering to be driven by directional similarity. We introduce $K$ learnable hyperspherical prototypes $\{\boldsymbol{\mu}_k\}_{k=1}^{K}$, each corresponding to a vMF component. 
The posterior assignment of node $i$ to cluster $k$ is computed by a temperature-scaled angular similarity:
\begin{equation}
q_{ik}=
\frac{
\exp\big(\kappa_i\boldsymbol{\mu}_k^\top \mathbf{z}_i / \beta\big)
}{
\sum_{j=1}^{K}
\exp\big(\kappa_i\boldsymbol{\mu}_j^\top \mathbf{z}_i / \beta\big)
},
\end{equation}
where $\kappa_i$ is the node-adaptive concentration and $\beta$ controls the assignment sharpness. The hyperspherical clustering objective is defined as
\begin{equation}
\mathcal{L}_{vMF}
=
-\frac{1}{n}
\sum\nolimits_{i=1}^{n}
\sum\nolimits_{k=1}^{K}
q_{ik}\kappa_i\boldsymbol{\mu}_k^\top\mathbf{z}_i
+
\beta
\sum\nolimits_{i=1}^{n}
\sum\nolimits_{k=1}^{K}
q_{ik}\log q_{ik}.
\end{equation}

Optimizing hyperspherical angular similarity enhances intra-cluster compactness and inter-cluster separability, mitigating imputation-induced representation collapse.



Although hyperspherical clustering improves separability, nodes near cluster boundaries may still receive ambiguous assignments. 
We therefore introduce a boundary-aware contrastive objective to focus on such uncertain nodes. 
The boundary sensitivity of node $v_i$ is measured by the normalized entropy of its posterior assignment:
\begin{equation}
B_i=-\frac{1}{\log K}\sum\nolimits_{k=1}^{K}q_{ik}\log(q_{ik}+\epsilon),
\end{equation}
where a larger $B_i$ indicates a more ambiguous boundary sample. 
We then define a node-adaptive concentration parameter
\begin{math}
\kappa_i=\frac{\kappa_0}{1+\gamma B_i},
\end{math}
which softens the assignments of boundary-sensitive nodes.

For each node, let $\boldsymbol{\mu}_{pos}$ and $\boldsymbol{\mu}_{neg}$ denote the prototypes of the most likely and second-most likely clusters. 
The boundary-aware contrastive loss is defined as
\begin{equation}
\mathcal{L}_{bound}
=
\sum\nolimits_{i=1}^{n}
B_i\cdot
\max\big(
0,
\Delta-\kappa_i
\big(
\boldsymbol{\mu}_{pos}^{\top}\mathbf{z}_i
-
\boldsymbol{\mu}_{neg}^{\top}\mathbf{z}_i
\big)
\big),
\end{equation}
where $\Delta$ is the angular margin. 
This loss emphasizes ambiguous boundary nodes and enlarges their angular margin between competing prototypes, leading to sharper cluster separation.

\subsection{Alternating Optimization}
We optimize RGCN in an alternating manner to avoid unstable joint updates. 
With the clustering variables fixed, we update the network parameters $\Theta$ by minimizing
\begin{equation}
\mathcal{L}_{total}
=
\mathcal{L}_{att}
+
\mathcal{L}_{str}
+
\lambda_1\mathcal{L}_{vMF}
+
\lambda_2\mathcal{L}_{bound}.
\end{equation}
Here, $\mathcal{L}_{att}$ and $\mathcal{L}_{str}$ correspond to attribute and structure reconstruction, while $\mathcal{L}_{vMF}$ and $\mathcal{L}_{bound}$ guide hyperspherical clustering and boundary enhancement. With $\Theta$ and the embeddings $\mathbf{Z}$ fixed, we then update the posterior assignments $q_{ik}$ and the prototypes $\boldsymbol{\mu}_k$ in an E/M-style clustering refinement. The assignments are updated according to the temperature-scaled angular similarity, and the prototypes are updated by the $\ell_2$-normalized concentration-weighted mean. 

\subsection{Efficiency Analysis and Scalability}

The computational complexity of RGCN is derived as follows. 
Under the Nyström approximation with $m=\lceil \sqrt{n} \rceil$, 
the overall per-epoch complexity is:

\begin{equation}
O_{\mathrm{total}} = O(N_t n^{\frac{3}{2}} d + |\mathcal{E}_{obs}| r + n r^2 + n K d).
\end{equation}

Compared with conventional methods requiring $O(n^2 d)$ dense kernel computation or $O(n^3)$ full-rank structure reconstruction, RGCN achieves sub-quadratic scalability with respect to the number of nodes. For a detailed step-by-step derivation, please refer to Appendix \ref{app:sec:efficiency}.

\begin{table*}[tb]
\centering
\setlength{\tabcolsep}{3pt}
\scalebox{0.72}{
 \begin{tabular}{lcccccccccccc}
\toprule[1pt]
\multirow{2}{*}{\textbf{Method}} &\multicolumn{4}{c}{\textbf{ACM}}&  \multicolumn{4}{c}{\textbf{REUT}} &\multicolumn{4}{c}{\textbf{HHAR}}  \\ \cmidrule(lr){2-5}\cmidrule(lr){6-9}\cmidrule(lr){10-13}
 & \textbf{ACC} & \textbf{NMI} & \textbf{ARI} & \textbf{F1}
 & \textbf{ACC} & \textbf{NMI} & \textbf{ARI} & \textbf{F1}
 & \textbf{ACC} & \textbf{NMI} & \textbf{ARI} & \textbf{F1} \\
\midrule
SDCN&43.1±0.3&10.6±0.4&1.5±0.4&38.5±1.4&
36.7±0.9&3.6±0.4&2.2±0.4&21.8±0.4&
\underline{43.4±1.0}&7.5±0.9&5.8±0.1&\underline{32.8±0.2}
\\
DFCN& 38.1±0.2&0.8±0.5&0.2±0.4&26.7±0.3&35.4±1.2&3.8±1.5&2.4±1.1&23.2±0.7
&28.5±7.7&16.7±15.7&6.4±11.0&16.0±9.1
\\
DCLN&40.5±1.4&4.3±1.2&5.6±1.1&35.4±1.0
&40.5±0.3&3.2±0.1&2.9±0.3&25.5±0.6
&34.3±0.9&30.3±2.0&23.1±2.0&23.5±1.4
\\
SCGC&41.6±1.7& 18.6±0.8& 19.5±0.6& 
22.3±1.2
&33.4±0.7&6.9±0.4&3.1±2.1&18.7±0.5
&35.4±1.0&35.5±2.2&25.8±2.4&22.7±1.6
\\
ABR&45.6±0.2&11.9±1.9&2.7±1.3&40.6±1.2
&41.4±0.8&7.7±1.2&4.0±1.6&29.9±1.3
&33.3±0.1&34.4±0.2&21.5±0.8&16.7±0.2

\\
 \midrule
AMGC&47.5±1.3&19.6±2.0&23.3±1.8&42.4±1.9
&42.6±1.7&2.5±0.2&5.3±0.2&16.4±2.3&36.2±0.8&46.0±1.0&30.9±1.0&21.6±1.8
\\
ITR&36.9±1.5&14.1±1.3&18.7±0.8&30.5±0.6
&43.1±0.5&3.1±0.1&1.2±0.2&13.0±0.3
&35.9±0.7&41.3±0.7&26.8±0.6&24.2±0.9
\\
AMMGC&49.6±0.7&21.2±1.3&25.6±2.0&40.7±0.8
&46.5±1.4&8.5±0.2&4.5±0.8&31.5±0.6
&37.7±0.7&37.5±0.5&25.4±0.3&27.0±0.2
\\
PMAGC&51.5±0.8&27.6±2.0&29.3±1.8&38.5±1.2
&42.7±0.6&9.7±0.1&6.0±0.3&34.2±0.8
&39.6±0.8&42.3±0.8&28.3±0.2&18.9±0.6

\\
CMVND&44.6±1.3&21.3±1.8&22.5±1.4&32.4±1.6
&\underline{47.1±1.2}&7.9±1.1&\underline{8.5±0.4}&34.2±0.8&43.2±1.0&\underline{54.0±1.4}&\textbf{32.9±1.1}&17.5±1.0
\\
RAM-MVC&\underline{53.2±1.2}&\underline{30.5±1.4}&\underline{32.6±0.9}&\underline{44.8±1.7}
&45.6±2.1&\underline{10.0±0.6}&7.0±2.1&\underline{35.0±1.9}
&19.0±1.1&0.5±1.8& 0.0±0.5&5.9±1.7
\\
\midrule
SMGCN & 33.6±1.3 & 3.9±1.0 & 5.7±0.5 & 24.8±1.2 & 43.1±1.5 & 3.1±1.7 & 0.0±0.0 & 14.2±1.9 & 31.1±1.2 & 20.5±1.2 & 11.8±1.5 & 10.7±0.8
\\ \midrule
\textbf{OURS}& \textbf{68.5±1.6}&\textbf{34.7±1.6}& \textbf{34.5±1.5}&\textbf{56.3±1.7} & \textbf{58.5±2.4}&\textbf{22.9±1.5}&\textbf{21.4±1.8}&\textbf{45.1±2.3}&\textbf{55.6±2.0}&\textbf{58.7±1.3}&\underline{32.6±1.9}&\textbf{46.7±2.2}

\\\midrule
&\multicolumn{4}{c}{\textbf{DBLP}}&  \multicolumn{4}{c}{\textbf{Co.CS}} &\multicolumn{4}{c}{\textbf{Arxiv}} \\
 \cmidrule(lr){2-5}\cmidrule(lr){6-9}\cmidrule(lr){10-13}
 & \textbf{ACC} & \textbf{NMI} & \textbf{ARI} & \textbf{F1}
 & \textbf{ACC} & \textbf{NMI} & \textbf{ARI} & \textbf{F1}
 & \textbf{ACC} & \textbf{NMI} & \textbf{ARI} & \textbf{F1} \\ \midrule


SDCN&35.8±1.4&0.4±0.0&0.2±0.3&21.2±0.7
&35.3±1.7&1.1±0.1&0.1±0.3&25.4±1.2
&16.1±1.6&0.0±0.1&0.0±0.0&0.7±0.0
\\
DFCN&39.0±1.6&2.3±0.4&0.2±0.4&25.5±2.2
&38.4±1.8&2.6±0.3&2.6±1.0&31.3±2.5
&18.7±0.3&0.1±0.0&0.5±0.0&0.3±0.0

\\
DCLN&30.0±1.5&0.1±0.0&0.1±0.0&14.0±0.2
&36.0±2.4&2.5±0.5&0.6±0.7&27.3±1.9
&23.4±0.3&0.9±0.0&0.2±0.1&5.6±0.4
\\
SCGC&31.4±0.3&2.4±0.2&1.3±0.1&16.7±0.9
&40.5±2.9&5.0±0.3&2.0±0.6&30.0±1.3 &
6.5±2.4&4.5±3.1&1.1±0.1&4.6±0.3
\\
ABR&43.1±0.0&0.1±0.0&3.0±0.2&15.1±0.0
&29.8±0.4&11.5±2.4&5.6±0.3&12.8±1.5
&16.9±1.0&1.0±0.2&2.1±0.3&2.5±0.2
\\
 \midrule
AMGC&\underline{46.9±1.0}&\underline{12.2±0.4}&\underline{10.0±0.6}&41.6±0.8
&41.0±2.6&0.6±0.2&-0.5±0.7&19.5±1.0
&23.8±1.2&7.6±1.8&4.9±1.7&1.9±1.4
\\
ITR&42.4±0.5&7.6±0.9&6.6±0.6&28.2±0.8
&40.4±1.3&23.9±0.5&14.6±1.4&23.8±1.3
&7.7±1.2 &7.0±0.5 &1.8±0.0 &5.4±0.4
\\
AMMGC&45.1±2.8&8.3±1.4&8.5±2.1&39.6±2.4
&\underline{53.1±0.2}&15.5±0.2&\underline{15.4±0.3}&\underline{31.6±0.1}
&21.4±1.3&7.8±0.6 &3.7±0.4 &13.1±1.5
\\
PMAGC
&45.2±2.1&7.4±0.6&8.9±0.7&\underline{42.1±1.6}
&49.6±0.3&16.1±0.4&11.3±0.4&30.7±0.2
&\underline{24.7±0.8}&\underline{9.6±1.2} &\underline{6.5±0.6} &\underline{14.4±1.0}
\\
CMVND&31.6±0.3&0.8±0.1&0.8±0.1&20.8±0.2
&33.2±0.8&\underline{24.3±1.2}&9.2±1.0&7.7±0.5
&16.1±0.0&0.0±0.0&0.0±0.0&0.9±0.0 \\
RAM-MVC&34.4±0.2&3.5±0.3&3.0±0.4&23.4±0.5
&25.3±1.4&7.2±1.7&1.6±1.5&8.1±1.6& OOM &OOM&OOM&OOM
\\
\midrule
SMGCN & 38.6±0.9 & 6.8±0.4 & 5.9±0.3 & 30.7±1.2 & 36.2±0.7 & 10.7±0.5  &8.6±1.2  & 18.1±0.9  & OOM &OOM&OOM&OOM
\\ \midrule


\textbf{OURS} & \textbf{55.3±1.4} & \textbf{16.8±1.5} & \textbf{13.4±1.1} & \textbf{45.9±2.0}
&\textbf{59.7±1.2}&\textbf{45.6±1.5}&\textbf{43.8±1.3}&\textbf{36.5±1.6} &\textbf{27.5±0.3}&\textbf{15.4±0.2} &\textbf{11.9±0.6 }&\textbf{17.1±0.8}

 \\
\bottomrule[1pt]
 \end{tabular}}
\caption{Performance comparison results (\%) under 50\% attribute and structure missing (expressed as mean ± standard deviation) under six datasets. \textbf{Bold} and \underline{underlined} figures represent the highest and second-highest performances, respectively. ``OOM'' represents memory limitation exceeded.}
\label{tab:main_result}
\end{table*}

\section{Experiments}
In this section, we conduct extensive experiments on multiple real-world datasets to validate the effectiveness of RGCN by answering the following research questions. 
\textbf{RQ1:} Does RGCN outperform SOTA DGC methods and DGC methods with node attribute imputation under the challenging scenario of simultaneous attribute and structure missingness? 
\textbf{RQ2:} How does each core component contribute to the overall performance? 
\textbf{RQ3:} How sensitive is RGCN to its key hyper-parameters? 
\textbf{RQ4:} How robust is RGCN under increasing levels of simultaneous attribute and structure missingness? 
\textbf{RQ5:} Can RGCN alleviate cross-stage error propagation compared with two-stage impute-then-cluster pipelines? 
\textbf{RQ6:} Does RGCN exhibit stable optimization and convergence behavior during training?

\subsection{Experimental Setup}
\label{sec:setup}

\textbf{Benchmark Datasets.}
To comprehensively evaluate the performance of RGCN, we conduct experiments on six widely-used real-world attributed graph datasets, covering citation networks, co-authorship networks, text data, and sensor-based activity records, including \textit{ACM}, \textit{REUT}, \textit{HHAR}, \textit{Co.CS}, \textit{DBLP}, and \textit{Arxiv}.
These datasets differ significantly in node scale, feature dimensionality, the number of clusters, and the balance of class distributions, thereby enabling a thorough examination of the generalization capability of our model. Detailed statistics of the benchmark datasets are provided in Appendix~\ref{app:sec:datasets}.


\textbf{Missing Value Simulation.}
To simulate realistic multi-view missingness, we construct two incomplete views for the benchmark datasets: (1) \textbf{Attribute Missingness}: we randomly mask entries of the node attribute matrix; (2) \textbf{Structure Missingness}: we randomly remove edges from the observed adjacency matrix. Unless otherwise specified, the default missing rate across all datasets is set to $r_X = r_A = 0.5$. To further evaluate the model's robustness under varying degrees of incompleteness, we specifically vary both $r_X$ and $r_A$ in the range of $[0.5, 0.8]$ on the ACM and REUT datasets.

\textbf{Baseline Methods.}
We compare RGCN against three categories of representative baselines: \textbf{Classical DGC methods}: SDCN \cite{Bo2020structuraldeep}, DFCN \cite{Tu2021deepfusion}, DCLN \cite{peng2023dualcontrastive}, SCGC \cite{liu2023simple}, ABR \cite{liu2025adaptivefeature}, \textbf{Attribute-missing graph clustering methods}: AMGC~\cite{tu2024attributemissinggraph},
ITR \cite{tu2022initializingthen}, AMMGC \cite{zhao2025attribute}, PMAGC \cite{guan2025prototype}, CMVND \cite{hu2025scalable}, and RAM-MVC \cite{hu2024reliable}. \textbf{Structure-missing graph clustering methods}: SMGCN \cite{hu2026structure}.

\textbf{Evaluation Metrics.}
Following the common practice in graph clustering, we adopt four standard evaluation metrics: clustering Accuracy (ACC), Normalized Mutual Information (NMI), Adjusted Rand Index (ARI), and F1 Score (F1).
Detailed definitions and mathematical formulations are provided in \textbf{Appendix}~\ref{app:sec:metrics}.


\textbf{Implementation Details.}
RGCN is implemented in PyTorch and optimized end-to-end using the Adam optimizer with a learning rate of $1\times 10^{-3}$ and a maximum number of training epochs $E = 200$. The embedding dimensionality $d$ is set to $10$, the low-rank dimensionality $r$ is set to $64$, the number of Nystr\"om landmark nodes is set to $m=\lceil\sqrt{n}\rceil$, which provides a practical trade-off between approximation quality and sub-quadratic computational cost. The number of ODE integration steps is $N_t = 10$, and the angular margin is $\Delta = 0.5$. The loss weights $\lambda_1$ and $\lambda_2$ are selected from $\{0.01, 0.1, 1.0\}$ via grid search, while their relative trade-off is further examined by varying $\lambda_1:\lambda_2$ in the sensitivity analysis. All experiments are conducted on a server equipped with an NVIDIA 4090 GPU. To alleviate the influence of randomness, each experiment is independently repeated 10 times, and both the mean and standard deviation are reported.

\begin{wraptable}[17]{r}{0.5\textwidth}
\vspace{-20pt} 
\centering
\caption{Clustering performance with 50\% missing rate under RGCN and its variants. \textbf{Bold} figures represent the highest performances for any column.}
\label{tab:ablation_results}

\setlength{\tabcolsep}{3pt}
\scalebox{0.82}{ 
\begin{tabular}{lccccc}
\toprule[1pt]
\textbf{Dataset} &Metric 
& \textbf{None} 
& w/o \textbf{DD} 
& w/o \textbf{MH} 
& \textbf{OURS} \\
\midrule

\multirow{3}{*}{ACM}
&ACC &43.4±1.2& 47.8±0.4  & 49.5±2.6 & \textbf{68.5±1.6} \\
&NMI &10.5±1.7& 13.4±1.8   & 14.3±1.4 & \textbf{34.7±1.6} \\
&ARI &12.3±1.0& 17.5±0.9   &  15.8±1.7 & \textbf{34.5±1.5} \\ \midrule

\multirow{3}{*}{REUT}
&ACC & 36.4±1.2 & 40.4±1.1 & 43.6±2.0 & \textbf{58.5±2.4} \\
&NMI & 5.6±0.8 & 7.5±1.3& 10.6±1.8 & \textbf{22.9±1.5} \\
&ARI & 2.9±0.3 & 11.3±0.9 & 11.5±1.3 & \textbf{21.4±1.8} \\\midrule

\multirow{3}{*}{Co.CS}&ACC & 13.6±0.2 & 36.1±0.3 & 40.2±0.8 &  \textbf{59.7±1.2} \\
&NMI & 1.4±0.0 & 40.4±0.4 & 16.8±0.6&\textbf{45.6±1.5}   \\
&ARI & 1.9±0.1 & 33.6±0.7 & 17.3±0.6&\textbf{43.8±1.3}   \\\midrule

\multirow{3}{*}{Arxiv}&ACC & 8.4±0.7 & 23.4±2.1 & 16.8±2.3 &\textbf{27.5±0.3}  \\
&NMI & 0.8±0.6 & 4.5±1.3 & 6.2±1.4 & \textbf{15.4±0.2} \\
&ARI & 5.4±1.1 & 3.7±2.5 & 5.1±2.0 & \textbf{11.9±0.6} \\
\bottomrule[1pt]
\end{tabular}}
\end{wraptable}

\subsection{Comparison Experiment}
\label{sec:comparison}


To answer \textbf{RQ1}, we compare RGCN with all baseline methods, with the overall results summarized in Table~\ref{tab:main_result}. From the reported results, we draw the following key observations: i) Compared to existing SOTA end-to-end clustering methods, our approach exhibits a significant performance advantage. This is primarily due to the inability of these methods to effectively handle missing data, which in turn introduces negative transfer signals that accumulate within the network, ultimately leading to a collapse in performance.
ii) Compared to methods that handle missing attributes in clustering, our approach still demonstrates superior performance. This is primarily attributed to the fact that once the missing graph structure is ignored, attribute imputation becomes unreliable, especially when key edges are missing. Incorrect propagation and imputation of attributes to the nodes lead to features that lack discriminative power, significantly degrading the clustering performance. 

\begin{figure}[t]
    \centering
    \includegraphics[width=1.0\linewidth, trim=8 10 5 20, clip]{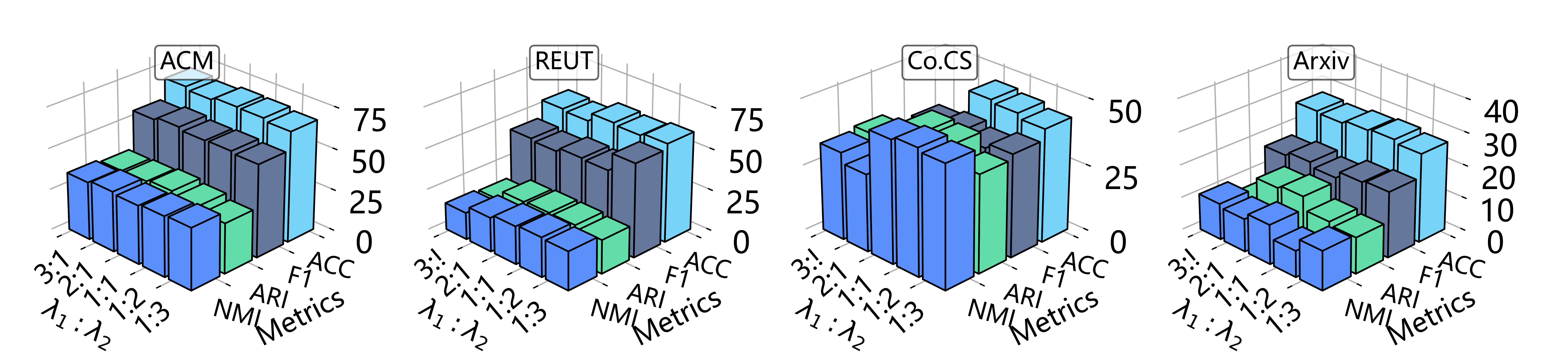}
    \caption{Sensitivity analysis results of RGCN under different $\lambda_1:\lambda_2$ ratios.}
    \label{fig:hyper-lambda}
\end{figure}

\subsection{Ablation Study}
\label{sec:ablation}

To answer \textbf{RQ2}, we design the following variants to evaluate the contribution of each key component of RGCN: \textbf{w/o DD} replaces the decoupled dual-branch alternative imputation with a unified joint imputation module; \textbf{w/o MH} removes the multi-hyperspherical mixture knowledge extraction; \textbf{None} denotes the removal of all the above components; and \textbf{OURS} denotes the complete RGCN. The experimental results are shown in Table \ref{tab:ablation_results}. From these, we can observe that the complete RGCN achieves optimal performance by establishing a principled collaboration between data reconstruction and cluster inference. Ablating the decoupled imputation strategy (\textbf{w/o DD}) severely exacerbates cross-view error propagation, causing noise to compound across incomplete modalities. Conversely, removing the multi-hyperspherical prior (\textbf{w/o MH}) leaves the objective susceptible to imputation bias, thereby inducing dimensional collapse and eroding inter-cluster margins. Naturally, the absence of both mechanisms (\textbf{None}) fails entirely to preserve the underlying semantic topology.

\subsection{Hyper-parameter Sensitivity Analysis}
\label{sec:sensitivity}

\paragraph{Impact of Loss Weights $\lambda_1$ and $\lambda_2$.}

\begin{wrapfigure}[10]{r}{0.55\textwidth}
\vspace{-15pt}
    \centering
    \includegraphics[width=\linewidth, trim=4 3 4 3, clip]{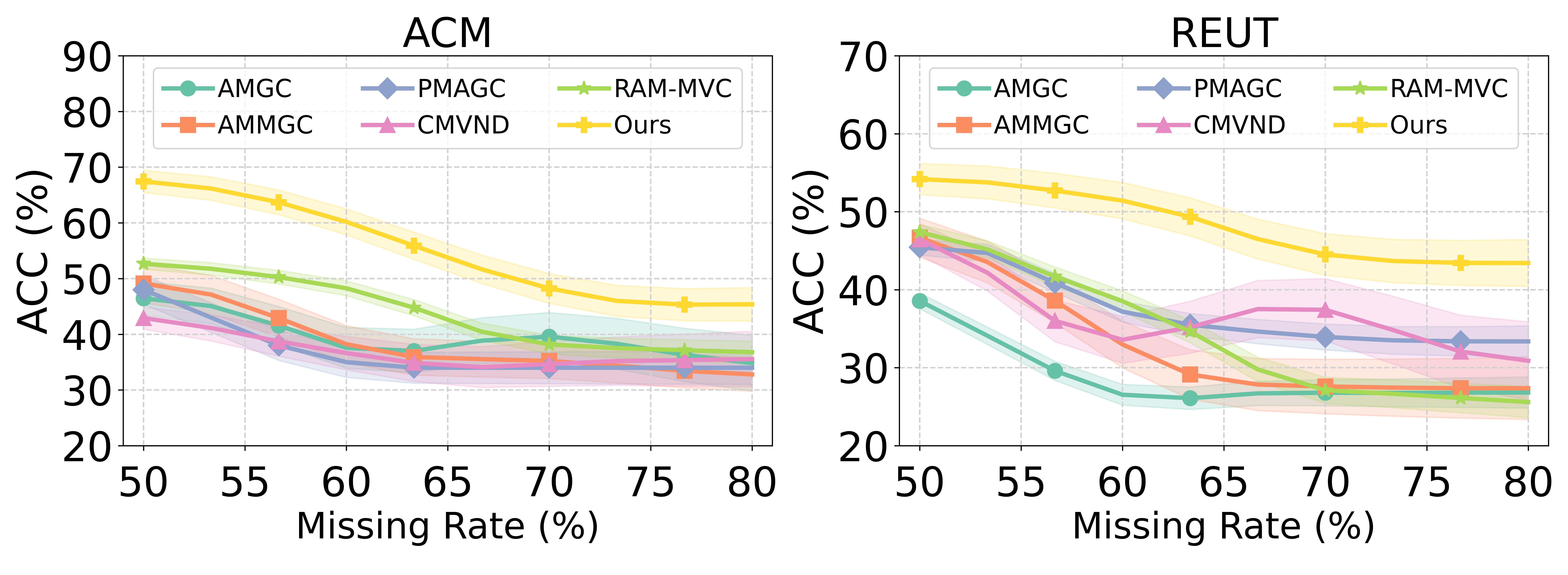}
    \caption{Clustering performance under missing rates from 50\% to 80\%, compared with six baselines.}
    \label{fig:generalization}
\vspace{-12pt}
\end{wrapfigure}



To answer \textbf{RQ3}, we study the sensitivity of RGCN to the trade-off between the multi-hyperspherical mixture prior and the contrastive boundary-enhancement objective, controlled by $\lambda_1$ and $\lambda_2$, respectively. We vary the ratio of $\lambda_1$ to $\lambda_2$ from $3:1$ to $1:3$. As shown in Fig.~\ref{fig:hyper-lambda}, RGCN maintains stable performance across different weight combinations. This indicates that a moderate balance between directional consistency and boundary sharpening can improve clustering robustness, whereas overly large weights may introduce excessive regularization and disturb the decoupled reconstruction process.

\paragraph{Robustness to Missing Rates.}


To answer \textbf{RQ4}, we evaluate the robustness of RGCN under increasing simultaneous attribute and structure missingness, with both $r_X$ and $r_A$ varied from $0.5$ to $0.8$. As shown in Fig.~\ref{fig:generalization}, RGCN consistently achieves the best performance and shows smoother degradation as the missing rate increases. These results indicate that: (i) existing methods that mainly address single-type incompleteness struggle to capture the dependency between attributes and structure under dual missingness; and (ii) the decoupled dual-pathway reconstruction in RGCN effectively reduces cross-view interference, enabling more reliable incomplete graph recovery.

\subsection{Comparison with Impute-then-Cluster Pipelines}
\label{sec:imputeCluster}

\begin{figure}[t]
    \centering
    \includegraphics[width=1.0\linewidth, trim=10 180 8 10, clip]{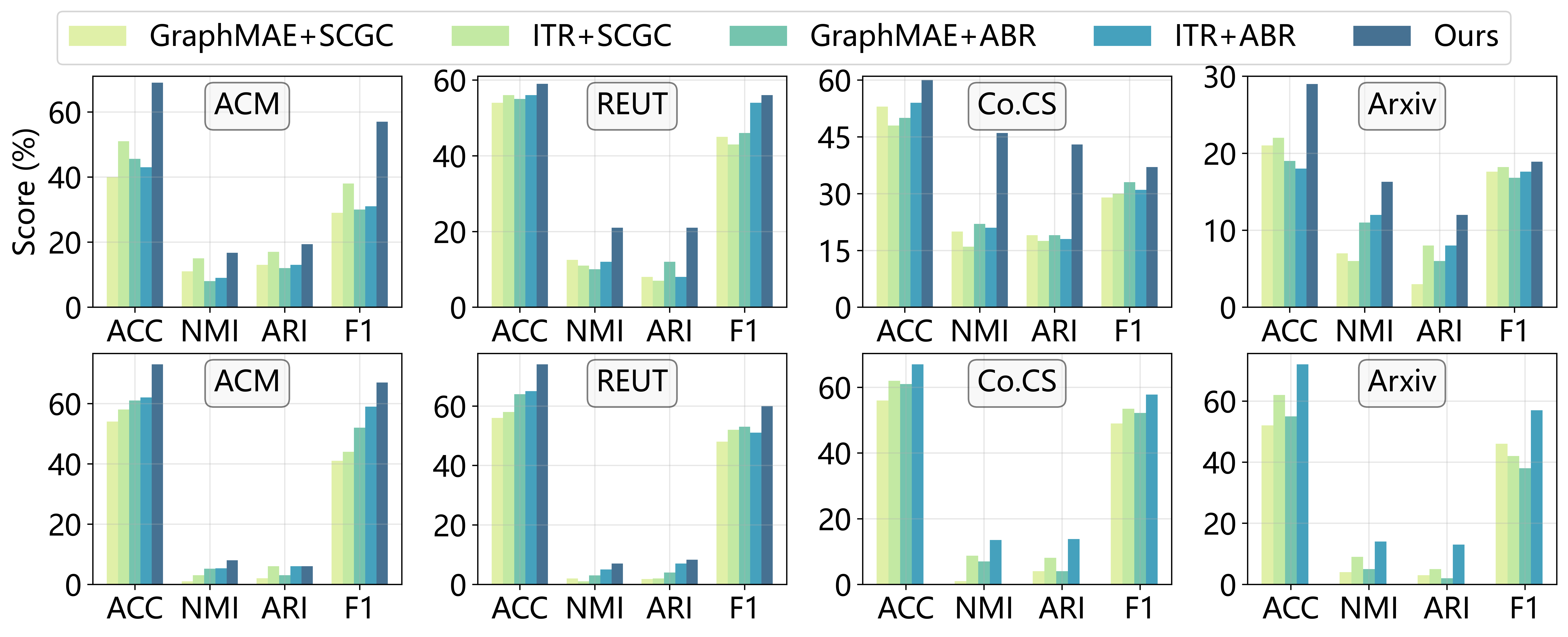}
    \caption{Comparison with two-stage impute-then-cluster pipelines.}
    \label{fig:impute_then_cluster}
\end{figure}

To answer \textbf{RQ5}, we evaluate cross-stage error propagation by comparing RGCN with four two-stage pipelines. As shown in Fig.~\ref{fig:impute_then_cluster}, RGCN consistently outperforms all baselines, yielding substantial ACC gains. This demonstrates that separating imputation from clustering severely amplifies reconstruction bias under dual-view missingness. Without explicit clustering guidance, two-stage methods suffer from global-mean drift and irreversible error propagation. In contrast, RGCN mitigates this issue by jointly optimizing decoupled reconstruction with a multi-hyperspherical prior, thereby preserving more discriminative representations.


\subsection{Model Convergence Analysis}
\begin{wrapfigure}[10]{r}{0.65\textwidth}
\vspace{-20pt}
    \centering
    \includegraphics[width=\linewidth, trim=70 2 70 0, clip]{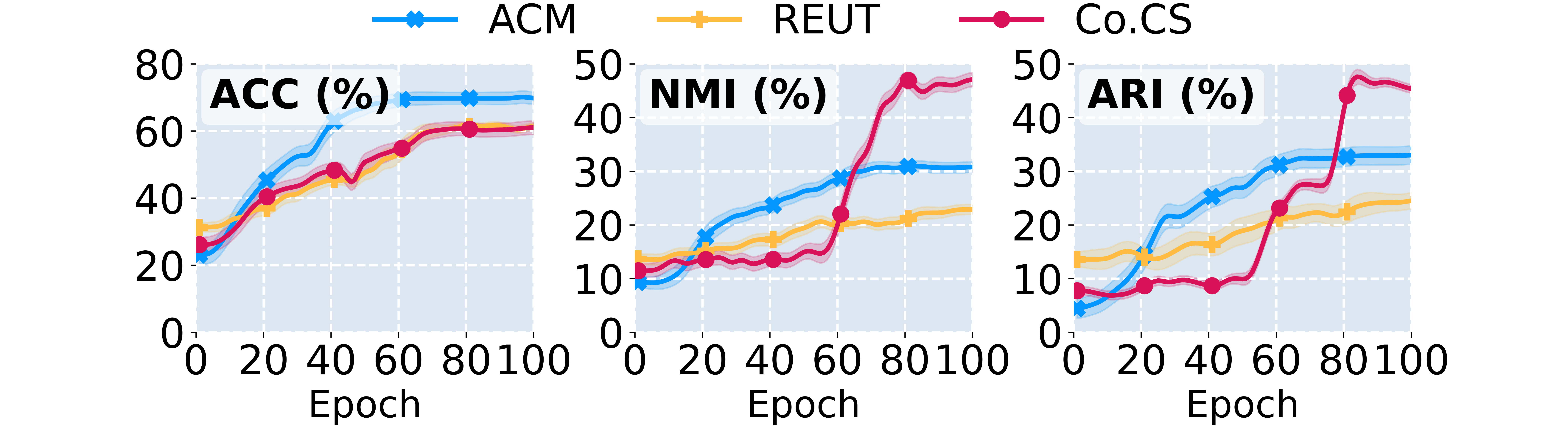}
    \caption{Epoch-wise ACC, NMI, and ARI performance curves on ACM, REUT, and Co.CS datasets.}
    \label{fig:convergence_epochs}
\vspace{-10pt}
\end{wrapfigure}

To answer \textbf{RQ6}, we examine the optimization behavior of RGCN by reporting the epoch-wise ACC, NMI, and ARI curves on three representative datasets. As shown in Fig.~\ref{fig:convergence_epochs}, the performance generally improves in the early training stage and then converges to stable plateaus. More importantly, the curves show no severe oscillations after convergence, suggesting that the alternating optimization between decoupled reconstruction and hyperspherical clustering is well coordinated.

\section{Conclusion}
In this paper, we propose a novel clustering network to solve the issue of multiple missing data in the graph. Compared to existing methods, our method can handle both missing node attributes and missing graph structures simultaneously, without requiring any label data. This dual-branch collaborative feature recovery provides a foundation for high-quality feature extraction, while multiple hyper-spheres boundary enhancement further optimizes the cluster structure. Extensive experiments on six benchmark datasets demonstrate that RGCN consistently outperforms its competitors. In future work, we plan to further improve the generalization ability of RGCN and extend it to more challenging and underexplored graph-level tasks.



\bibliography{main}
\bibliographystyle{unsrt}
\appendix
\section{Preliminaries and Notations}
\label{app:Notations}

Suppose that $\mathcal{G}=(\mathcal{V},\mathbf{X},\mathbf{A})$ denotes a complete attributed graph with $n$ nodes, where $\mathcal{V}$ is the node set, $\mathbf{X}\in\mathbb{R}^{n\times d}$ is the complete node attribute matrix, and $\mathbf{A}\in\mathbb{R}^{n\times n}$ is the complete topological adjacency matrix. To model simultaneous attribute and structure missingness, we introduce two binary observation masks, namely the attribute mask $\mathbf{M}_X\in\{0,1\}^{n\times d}$ and the structure mask $\mathbf{M}_A\in\{0,1\}^{n\times n}$, where an entry equal to $1$ indicates that the corresponding value is observed and an entry equal to $0$ indicates that it is missing. Accordingly, the observed incomplete attribute matrix and adjacency matrix are defined as
\begin{equation}
    \mathbf{X}_{obs}=\mathbf{M}_X\odot\mathbf{X},
    \qquad
    \mathbf{A}_{obs}=\mathbf{M}_A\odot\mathbf{A},
\end{equation}
respectively.

To provide valid self-supervised training signals for attribute imputation, we further introduce a denoising mask $\mathbf{M}_D\in\{0,1\}^{n\times d}$ during training. The denoising mask is sampled only from the observed attribute entries and satisfies $\mathbf{M}_D\leq \mathbf{M}_X$, where $\mathbf{M}_D(i,j)=1$ indicates that the corresponding observed attribute is deliberately hidden and used as a reconstruction target. Based on this mask, the corrupted attribute input is defined as
\begin{equation}
    \tilde{\mathbf{X}}_{obs}
    =
    (\mathbf{M}_X-\mathbf{M}_D)\odot\mathbf{X}_{obs}.
\end{equation}

In the attribute inference branch, $\mathbf{X}_{pred}$ denotes the full attribute matrix predicted by the diffusion-based attribute predictor, while $\hat{\mathbf{X}}$ denotes the final completed attribute matrix used for subsequent structure inference and clustering. Specifically, the final completed attribute matrix preserves the originally observed entries and fills only the truly missing entries with the predicted values:
\begin{equation}
    \hat{\mathbf{X}}
    =
    \mathbf{M}_X\odot\mathbf{X}_{obs}
    +
    (\mathbf{1}-\mathbf{M}_X)\odot\mathbf{X}_{pred}.
\end{equation}
The recovered adjacency matrix is denoted by $\hat{\mathbf{A}}$. In the structure inference branch, we consistently use $\mathbf{E}_s$ to denote the low-rank structural embedding used for reconstructing $\hat{\mathbf{A}}$.

Unless otherwise specified, the attribute reconstruction loss is computed on the deliberately masked observed entries indicated by $\mathbf{M}_D$, while the structure reconstruction loss is computed on the observed structural entries indicated by $\mathbf{M}_A$.

To ensure clarity, the primary data symbols and variables used throughout this paper are summarized in Table \ref{tab:notations}.

\begin{table}[htbp]
\centering
\caption{Summary of Key Notations in RGCN}
\label{tab:notations}
\begin{tabular*}{\textwidth}{@{\extracolsep{\fill}}clcl@{}}
\toprule
\textbf{Symbol} & \textbf{Description} & \textbf{Symbol} & \textbf{Description} \\
\midrule
$\mathcal{G}$ & Attributed graph & $\mathcal{V}$ & Node set \\
$n$ & Node count & $d$ & Feature dimension \\
$\mathbf{X}$ & Complete attribute matrix & $\mathbf{A}$ & Complete adjacency matrix \\
$K$ & Cluster count & $r$ & Rank dimension \\
$m$ & Landmark size & $\mathbf{X}_{obs}$ & Observed attribute matrix \\
$\mathbf{A}_{obs}$ & Observed adjacency matrix & $\mathbf{M}_X$ & Binary mask for attributes \\
$\mathbf{M}_A$ & Binary mask for structure & $\mathbf{M}_D$ & Denoising mask for attributes \\
$\tilde{\mathbf{X}}_{obs}$ & Corrupted observed attribute matrix & $\mathbf{X}_{pred}$ & Predicted full attribute matrix \\
$\hat{\mathbf{X}}$ & Final completed attribute matrix & $\hat{\mathbf{A}}$ & Recovered structure \\
$\mathbf{Z}$ & Joint representation matrix & $\theta$ & Encoder parameters \\
$\mathbf{H}(s)$ & Latent state matrix during diffusion & $\mathbf{W}_v$ & Propagation rate weight matrix \\
$\Delta_X(s)$ & Dynamic feature-space Laplacian & $\mathbf{E}_s$ & Low-rank structural embedding \\
$\mathbf{L}_{obs}$ & Laplacian of observed adjacency & $\boldsymbol{\mu}_k$ & Prototype for cluster $k$ \\
$\kappa_i$ & Node-adaptive concentration & $q_{ik}$ & vMF posterior assignment probability \\
$B_i$ & Node-wise boundary sensitivity & $\lambda_i$ & Trade-off hyperparameters \\
\bottomrule
\end{tabular*}
\end{table}
\section{Efficiency Analysis and Scalability}
\label{app:sec:efficiency}

Standard implementations of continuous graph diffusion and exact low-rank optimization usually suffer from quadratic or cubic computational costs, limiting their applicability to large-scale graphs. RGCN reduces the computational overhead through Nystr\"om-based feature-affinity approximation and low-rank structural reconstruction.

\paragraph{Attribute diffusion complexity.}
The exact computation of the dynamic feature-affinity kernel requires $O(n^2d)$ operations. To avoid constructing the full dense kernel, we adopt the Nystr\"om approximation by sampling $m$ landmark nodes $(m\ll n)$. With $N_t$ Euler discretization steps, the time complexity of attribute diffusion is:
\begin{equation}
O_{\mathrm{att}} = O(N_t n m d).
\end{equation}
In our implementation, we set $m=\lceil\sqrt{n}\rceil$ to balance approximation quality and efficiency. Therefore, the attribute diffusion complexity becomes:
\begin{equation}
O_{\mathrm{att}} = O(N_t n^{\frac{3}{2}} d),
\end{equation}
which is sub-quadratic in $n$ and lower than the exact $O(n^2d)$ dense-kernel computation.

\paragraph{Structural reconstruction complexity.}
For structural topology recovery, conventional full-rank matrix reconstruction or SVD-based optimization may incur $O(n^3)$ complexity. RGCN avoids this cost by parameterizing the structural logits as a low-rank inner product $\mathbf{E}_s\mathbf{E}_s^\top$, where $\mathbf{E}_s\in\mathbb{R}^{n\times r}$ and $r\ll n$. This gives the structural reconstruction overhead:
\begin{equation}
O_{\mathrm{str}} = O(|\mathcal{E}_{obs}|r + nr^2).
\end{equation}

\paragraph{Overall complexity.}
The vMF posterior computation and boundary-aware contrastive objective require $O(nKd)$ operations. Therefore, under $m=\lceil\sqrt{n}\rceil$, the total per-epoch complexity of RGCN is:
\begin{equation}
O_{\mathrm{total}} = O(N_t n^{\frac{3}{2}}d + |\mathcal{E}_{obs}|r + nr^2 + nKd).
\end{equation}
This complexity is not strictly linear in $n$, but it remains sub-quadratic with respect to the diffusion component and avoids both dense-kernel construction and cubic matrix-decomposition costs.
\section{Details of Benchmark Datasets}
\label{app:sec:datasets}
To comprehensively evaluate the generalization capability and robustness of the proposed RGCN framework under complex missing scenarios, we conduct experiments on six widely used benchmark datasets, as summarized in Table \ref{II}. These datasets are strategically selected to cover a diverse spectrum of data domains, topological structures, and graph scales.

\textbf{Citation Networks (Natural Graphs):} ACM \cite{tang2008arnetminer}, DBLP \cite{tang2008arnetminer}, and Arxiv \cite{hu2020open} are standard benchmarks in academic citation networks. In these networks, nodes represent academic papers and edges denote citation relationships. While ACM and DBLP serve as typical small-to-medium scale evaluations, Arxiv is a large-scale dataset encompassing nearly 170,000 nodes and over a million edges. The inclusion of Arxiv is crucial, as it provides strong empirical evidence for the improved scalability and efficiency of our framework on large-scale attributed graphs. In addition, Co.Cs \cite{shchur2018pitfalls} is a co-authorship graph, where nodes correspond to authors and edges indicate collaboration relationships, making it another widely used benchmark for node-level graph learning tasks.

\textbf{Non-Graph Domains (Constructed Graphs):} To further verify the topological adaptability of RGCN beyond highly homophilous citation networks, we include REUT \cite{lewis2004rcv1} and HHAR \cite{stisen2015smart}. These are originally non-graph datasets from the news classification and human activity recognition domains, respectively. By constructing edges based on the k-nearest neighbors (k-NN) similarity of their text or sensor feature representations, we transform them into graph-structured data. Evaluating on these datasets demonstrates the model's robustness in handling diverse feature spaces and artificially constructed topologies, proving that our decoupled imputation method is not overfitted to a specific type of natural graph structure.

\begin{table*}[ht]
    \centering
    \label{app:sec:benchmark}
    \caption{Datasets summary.}
    \begin{tabular}{ccccccc}
        \toprule
        \textbf{Dataset} &\textbf{Type} & \textbf{Dimensions} &\textbf{Samples} & \textbf{Edges} & \textbf{Clusters} & \textbf{Domain}  \\
        \midrule
        ACM        & Graph &  1870   & 3025 & 13128 & 3 & Citation\\
        REUT          & Text &  2000  & 10000  & 30000 & 4 & News\\
        HHAR & Record & 561  & 10299  & 30897 & 6 & Human Activity \\
        DBLP    & Graph & 334  & 4057  & 3528 & 4 & Citation\\
        Arxiv       & Textural & 128  & 169343  & 1166243 & 40 & Citation\\
        Co.CS       & Graph & 6805  & 18333  & 81894 & 15 & Co-authorship\\
        \bottomrule
    \end{tabular}
    \label{II}
\end{table*}



\section{Evaluation Metrics}
\label{app:sec:metrics}

In this study, we evaluate the performance of our proposed method using four standard metrics to comprehensively assess different aspects of clustering quality and model effectiveness:

\begin{enumerate}
    \item \textbf{Clustering Accuracy (ACC):} This metric evaluates the overall correctness of the clustering assignments by comparing the predicted cluster labels with the ground truth.
    \begin{equation}
    \text{ACC} = \frac{1}{N} \sum_{i=1}^N \mathbb{I}(y_i = \hat{y}_i)
    \end{equation}
    where $y_i$ and $\hat{y}_i$ denote the true and predicted labels for the $i$-th sample, respectively, and $\mathbb{I}(\cdot)$ is the indicator function.

    \item \textbf{Normalized Mutual Information (NMI):} NMI quantifies the amount of statistical information shared between the true and predicted label distributions, measuring their consistency.
    \begin{equation}
    \text{NMI} = \frac{I(X, Y)}{\sqrt{H(X) H(Y)}}
    \end{equation}
    where $I(X, Y)$ is the mutual information between the true labels $X$ and predicted labels $Y$, while $H(X)$ and $H(Y)$ represent their respective entropies.

    \item \textbf{Adjusted Rand Index (ARI):} ARI measures the similarity between the true and predicted label assignments by considering all possible pairwise comparisons, while rigorously adjusting for chance.
    \begin{equation}
    \text{ARI} = \frac{\text{RI} - \mathbb{E}[\text{RI}]}{\max(\text{RI}) - \mathbb{E}[\text{RI}]}
    \end{equation}
    where $\text{RI}$ is the Rand Index, and $\mathbb{E}[\text{RI}]$ denotes the expected value of the Rand Index under random label assignments.

    \item \textbf{F1-Score (F1):} As the harmonic mean of precision and recall, the F1-Score provides a balanced evaluation of the model's ability to correctly identify pairwise clustering relationships.
    \begin{equation}
    \text{F1} = \frac{2 \cdot \text{Precision} \cdot \text{Recall}}{\text{Precision} + \text{Recall}}
    \end{equation}
    where precision and recall are calculated based on the true positive, false positive, and false negative clustering pairs.
\end{enumerate}

These metrics provide a comprehensive evaluation of our method, capturing both the quality of the clustering results and their alignment with the underlying data structure. Detailed performance results and comparisons against existing baselines are presented in the Comparative Experiments section.
\paragraph{Limitations.}
Despite its effectiveness, RGCN still has several limitations. First, although the decoupled dual-branch reconstruction reduces cross-view interference, the alternating optimization may introduce additional computational cost on large-scale graphs with severe attribute and structure missingness. Second, RGCN is mainly designed for static attributed graphs, while its extension to dynamic, heterogeneous, or multi-view graph data remains to be explored. Third, the current framework follows a centralized setting, which may limit its applicability to privacy-sensitive scenarios where incomplete graph data are distributed across multiple clients. In future work, we will further improve the scalability and adaptability of RGCN and extend it to federated incomplete graph clustering scenarios.

\section{Broader Impacts}
This work studies general graph representation learning and clustering methods, without targeting specific applications or deployment scenarios. Its impact is mainly positive in broad data analysis settings, and no societal risks are expected.



\end{document}